\documentclass[runningheads]{llncs}

\usepackage{eccv}

\usepackage{eccvabbrv}

\usepackage{graphicx}
\usepackage{booktabs}

\usepackage{amsmath,amssymb}
\usepackage{listings}

\usepackage[accsupp]{axessibility}  %

\newcommand{\x}{\mathbf{x}}
\newcommand{\rtoken}{\mathbf{r}}
\newcommand{\xadv}{\mathbf{x}^{\text{adv}}}
\newcommand{\linv}{\mathcal{L}_{\text{inv}}}
\newcommand{\ladv}{\mathcal{L}_{\text{adv}}}
\newcommand{\loss}{\mathcal{L}}
\newcommand{\ltwo}[1]{\lVert#1\rVert}

\newcommand{\linvformula}{ \frac{f([\rtoken, \x]) \cdot f(\x)}{\ltwo{f([\rtoken, \x])} \ltwo{f(\x)}}  }

\newcommand{\ladvformula}{ \frac{f([\rtoken, \xadv]) \cdot f(\x)}{\ltwo{f([\rtoken, \xadv])} \ltwo{f(\x)}}  }

\usepackage{hyperref}

\usepackage{orcidlink}
\usepackage[absolute]{textpos}

\begin{document}

\definecolor{ourgray}{rgb}{0.5, 0.5, 0.5}
\newcommand{\darkgrayed}[1]{\textcolor{ourgray}{#1}}
\begin{textblock}{8}(4, 0.7)
\begin{center}
\darkgrayed{This paper has been accepted for publication at the \\
European Conference on Computer Vision (ECCV), 2024}
\end{center}
\end{textblock}

\title{Robustness Tokens: Towards Adversarial Robustness of Transformers} 

\author{Brian Pulfer\orcidlink{0000-0003-0809-6978} \and
Yury Belousov\orcidlink{0000-0001-6461-734X} \and
Slava Voloshynovskiy\orcidlink{0000-0003-0416-9674}}

\authorrunning{B.Pulfer et al.}

\institute{University of Geneva, Department of Computer Science, Switzerland
\email{\{Brian.Pulfer,Yury.Belousov,svolos\}@unige.ch}
}

\maketitle

\begin{abstract}
  Recently, large pre-trained foundation models have become widely adopted by machine learning practitioners for a multitude of tasks. Given that such models are publicly available, relying on their use as backbone models for downstream tasks might result in high vulnerability to adversarial attacks crafted with the same public model. In this work, we propose Robustness Tokens, a novel approach specific to the transformer architecture that fine-tunes a few additional private tokens with low computational requirements instead of tuning model parameters as done in traditional adversarial training. We show that Robustness Tokens make Vision Transformer models significantly more robust to white-box adversarial attacks while also retaining the original downstream performances.
\end{abstract}

\begin{figure}[ht]
    \centering
    \includegraphics[width=\linewidth]{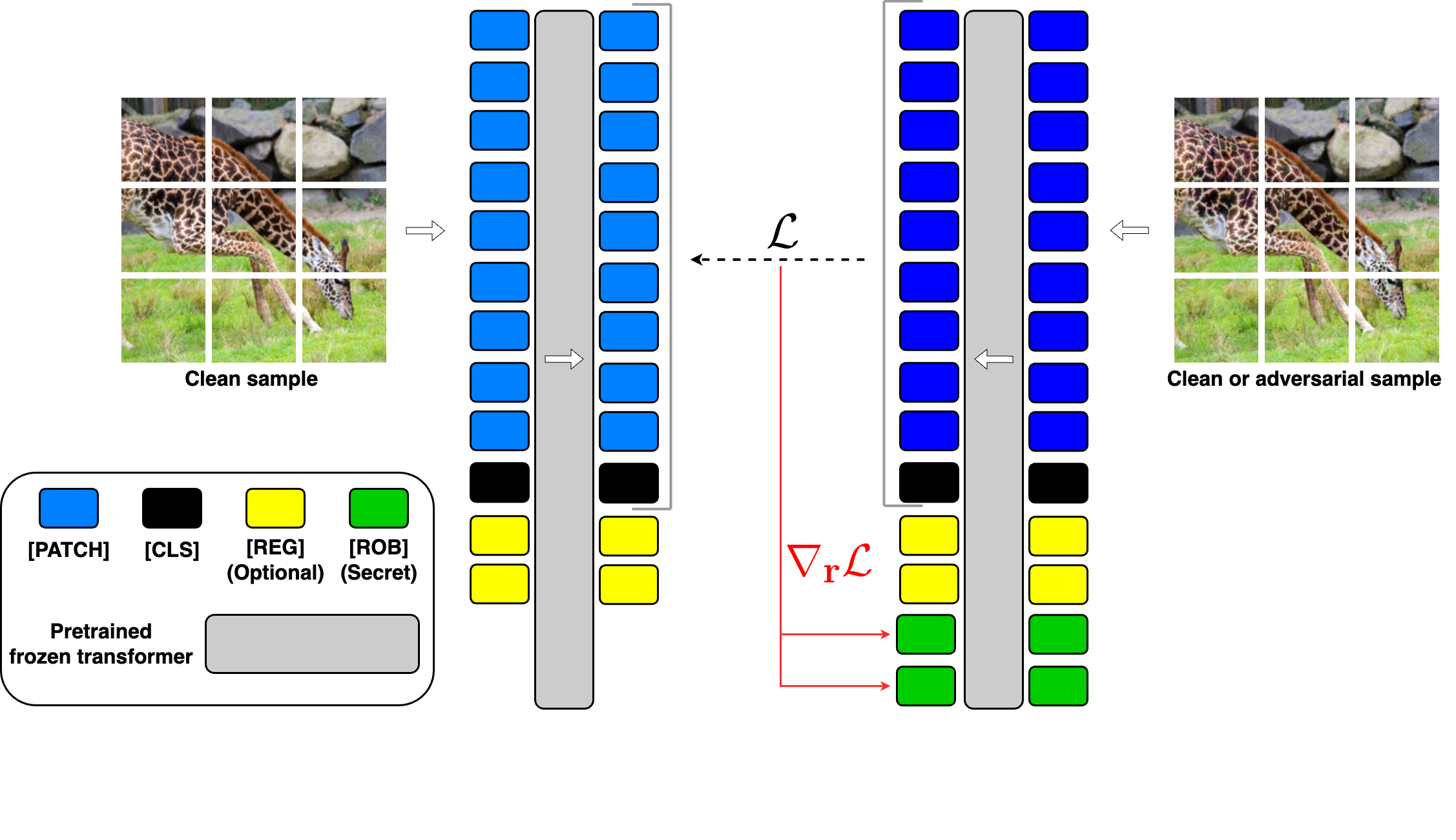}
    \caption{Schematic representation of Robustness Tokens}
    \label{fig:training}
\end{figure}

\section{Introduction}
\label{sec:intro}
Most foundation models nowadays are based on the transformer architecture \cite{transformer, vit}. The high computational cost of pre-training such models on a large amount of data, which might not even be publicly available, discourages practitioners from pre-training custom models and instead encourages using publicly available foundation models as backbones and fine-tuning task-specific heads and parameters at a much lower computational budget \cite{lora, qlora}. However, this trend could represent an advantage for malicious attackers, who might make reasonable guesses on the architecture type and initial weights of the model being attacked. By using the publicly available pre-trained model as a surrogate \cite{ban2022pre, inkawhich2023adversarial}, adversarial attacks could have a high impact even on the fine-tuned model without a proper defense mechanism given the high transferability between surrogate and victim \cite{sitawarin2023defending}.

Recent work \cite{attnsinks, registers, li2021prefix, sun2024massive} has shown that additional tokens that do not depend on the input sequence can be beneficial to the performances of transformers. To date, to the best of our knowledge, no work has studied the potential benefit of additional tokens in terms of adversarial robustness. Motivated by recent findings, we propose Robustness Tokens (\texttt{ROB} tokens), additional learnable tokens that are appended to the input sequence to make transformer models more robust to adversarial attacks while retaining the original feature extraction capabilities. Differently from \cite{registers}, we propose to not re-train pre-existing foundation models from scratch, but to only learn a few additional tokens. Hence, Robustness Tokens can be efficiently trained by practitioners and kept secret, adding a layer of uncertainty for malicious attackers. We further show that a few Robustness Tokens, while being relatively few extra parameters, are sufficient to drastically drop the efficacy of white-box attacks.

While our methodology is general to the transformer model and its variations \cite{vit, mann2020language, devlin2018bert, whisper, liu2021swin, arnab2021vivit, xie2021segformer, touvron2022deit}, and more broadly to sequence models \cite{hochreiter1997long, peng2023rwkv, sun2023retentive, gu2023mamba, hatamizadeh2024vir, zhu2024visionmamba}, in this work we focus on the application of Robustness Tokens in the domain of computer vision. We extensively explore the efficacy of Robustness Tokens when applied to the DiNOv2 \cite{dinov2} model, while also showing their applicability to DEIT-III \cite{touvron2022deit} and OpenCLIP \cite{opencliprepo} models. We demonstrate that Robustness Tokens improve robustness both in feature space, classification, and segmentation tasks, generalize across adversarial attacks, and can be trained at little computational cost.

\section{Related work}
\label{sec:related}

\subsection{Vision Foundation Models}
Foundation models, based on the transformer architecture and trained on extensive datasets, pose significant challenges due to the vast amount of data, computational demands, environmental impact, and logistical complexities associated with training on hardware platforms \cite{awais2023foundational, bommasani2021opportunities, yuan2022decentralized, zhou2024training}. Consequently, practitioners are increasingly opting to fine-tune publicly available pre-trained models on their specific data, leveraging the ease and efficiency of this approach compared to the arduous task of training models from scratch \cite{lora, qlora, he2021towards, lian2024less, jiang2024supervised}. This trend is likely to persist as larger models consistently yield improved performance, thereby reinforcing the preference for utilizing pre-trained backbones in practical applications \cite{hestness2017deep, bahri2021explaining}.

In the field of computer vision, Vision Foundation Models (VFMs) typically consist of Vision Transformers (ViTs) \cite{vit} models pre-trained with Self-Supervised Learning (SSL) methods that leverage self-distillation, masked image modeling, contrastive learning and more techniques \cite{balestriero2023cookbook}. Popular publicly available models include DiNOv2 \cite{dinov2}, iBOT \cite{ibot}, I-JEPA \cite{ijepa}, CLIP \cite{radford2021learning} and its variations \cite{blip, flip, glip, align} and many more \cite{beit2, cae, mae, msn, aim}.

\subsection{Adversarial Attacks and Adversarial Training}

Neural networks are known to be vulnerable to adversarial attacks \cite{nguyen2015deep, demontis2019adversarial}. Gradient-based methods such as FGSM \cite{goodfellow2014explaining} and PGD \cite{pgd} and optimization-based methods like DeepFool \cite{moosavi2016deepfool} and CW \cite{carlini2017towards} are extremely effective, especially in the white-box setting where the assumption is that the attacker has full knowledge of the model architecture, parameters, and training data used by the defender. While it is believed that SSL methods can improve model robustness \cite{NEURIPS2019_a2b15837}, recent work has shown that VFMs are still very vulnerable to adversarial attacks \cite{Fort2021CLIPadversarial, inkawhich2023adversarial, ban2022pre, Schlarmann_2023_ICCV, rando2022exploringadversarialattacksdefenses}.

To mitigate the devastating effects of adversarial attacks, adversarial training proposes to use adversarial samples in the training process to robustify the obtained model by changing the model parameters such that all samples are classified correctly \cite{bai2021recent}.

In this work, we propose a novel adversarial training technique that does not modify the model parameters but rather only modifies the input to the sequence model instead by adding additional sequence elements. This design results in tuning very few parameters compared to those of a VFM and is advantageous both in terms of optimization and compute requirements.

\subsection{Additional Tokens}

Recent work has shown that additional learned tokens that do not depend on the input sequence can be beneficial to the performances of transformers.

In particular, \cite{registers} proposed "register" tokens, additional learnable tokens used for the pre-training of DiNOv2 \cite{dinov2} models to enhance performances while removing artifacts from attention maps.

Similarly, in Natural Language Processing, \cite{attnsinks} prepend learnable tokens, called "attention sinks", which are responsible for storing the context of the sequence and enable transformer models to generate high-quality outputs despite extremely long context sizes.

Similarly to our work, Prefix-tuning \cite{li2021prefix} only learns a few additional tokens instead of fine-tuning all the parameters of a language model. The authors find that for low data settings, tuning the prefix of the transformer model yields better results and extrapolation than full fine-tuning, which is particularly relevant for most practitioners fine-tuning a foundation model.

Differently from this work, however, the goal of register tokens, attention sinks, and prefix tokens trained with Prefix-tuning is to obtain better feature extraction capabilities or performances on downstream tasks, whereas our objective is to improve the adversarial robustness of features extracted with foundation models, effectively defending from a multitude of attacks.

Most recently, \cite{sun2024massive} observed "massive activations" in the hidden states of decoder and encoder transformers, and found that additional tokens, such as registers and attention sinks, become responsible for carrying high absolute norm activations that seem to act as bias terms for the computation of attention, as previously hinted in \cite{off_by_one}.

\section{Adversarial Training with Robustness Tokens}
\label{sec:main}
Our goal is to append secret additional tokens, named Robustness Tokens, such that two conditions are met: a) the addition of Robustness Tokens to clean samples changes as least as possible the obtained feature representation and b) adding Robustness Tokens to adversarial samples results in extracted features that are as close as possible to those extracted with the original model on the clean samples. The first condition is needed to keep good feature representations in the common case the sample is not an adversary. The second condition robustifies the model against samples that take advantage of the knowledge and activations of the original model. We call these conditions invariance and adversary respectively.

To meet the invariance condition, we simply maximize the cosine similarities of features extracted by the transformer with and without the use of Robustness Tokens on clean samples:

\begin{equation}
    \linv(\rtoken) = \mathop{\mathbb{E}}_{\x \sim p_{\text{data}}} \left[ \linvformula \right]
\end{equation}

where $p_{\text{data}}$ is the data distribution, $f$ denotes the pre-trained model, $\x$ are the tokens of the clean sample sequence, $\rtoken$ are the secret Robustness Tokens appended to the input sequence, $[ \cdot ]$ denotes the concatenation operation across the sequence dimension.

To meet the adversary condition, we once more enforce the constraint by maximizing the cosine similarity of the obtained feature representations, this time aligning the representation for an adversarial sample instead:

\begin{equation}
    \ladv(\rtoken) = \mathop{\mathbb{E}}_{\x \sim p_{\text{data}}} \left[ \ladvformula \right],
\end{equation}

where the adversarial sample $\xadv = A(f, \x)$ is obtained by running a task-agnostic adversarial attack $A$ on the original model (without the use of Robustness Tokens) with the clean sample $\x$. Any kind of task-agnostic adversarial attack might be used to obtain $\xadv$, including black box or white box, and any combination of such attacks.

Our total objective function is the sum of the invariance and adversary terms, which we aim to maximize:

\begin{equation}
    \label{eq:3}
    \loss(\rtoken) = \linv(\rtoken) + \ladv(\rtoken)
\end{equation}

\begin{figure}[ht]
    \centering
    \includegraphics[width=\linewidth]{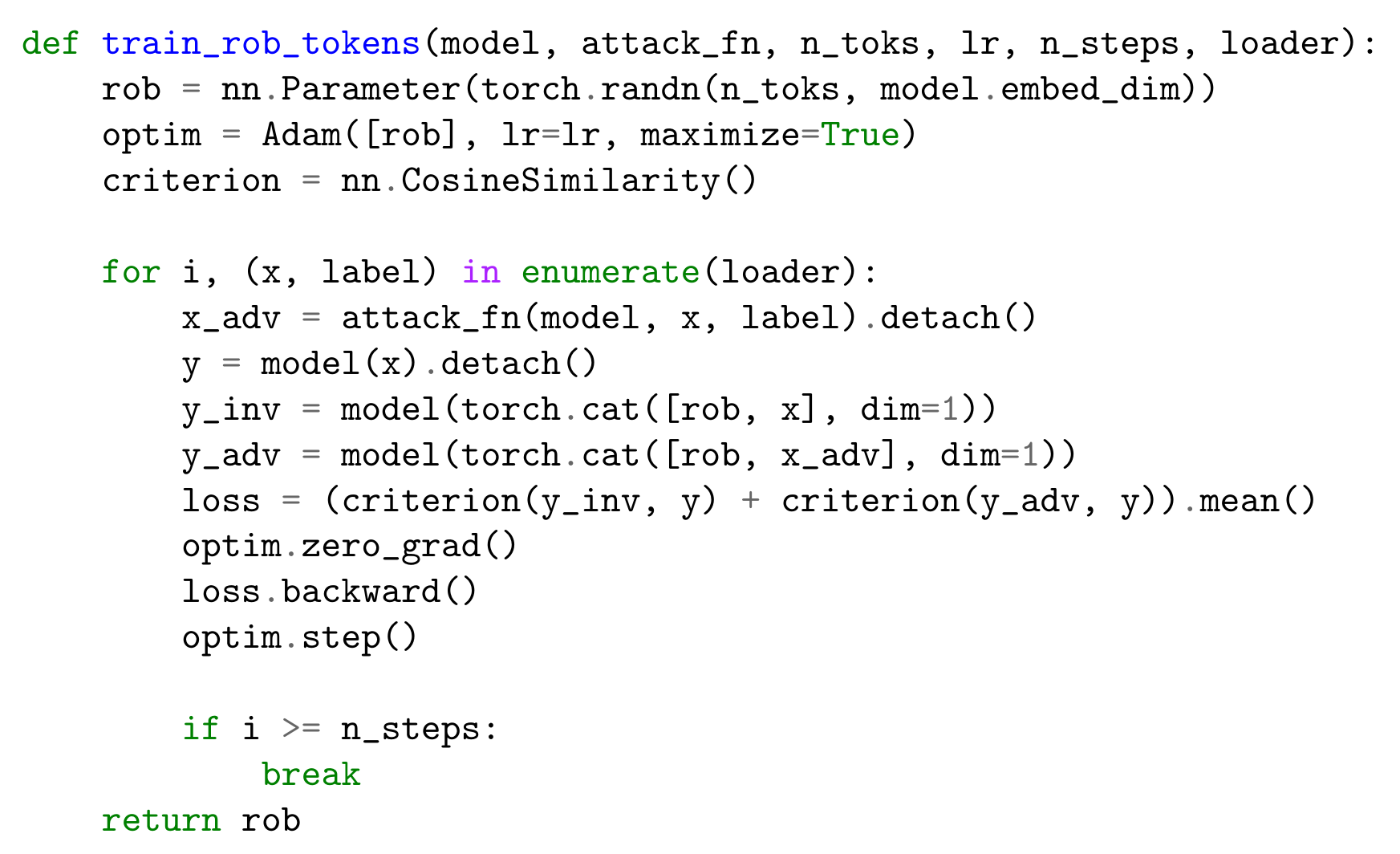}
    \caption{Pytorch-like pseudocode for training Robustness Tokens.}
    \label{fig:pseudocode}
\end{figure}

Notice that we optimize \cref{eq:3} only with respect to the newly introduced Robustness Tokens $\rtoken$, keeping the model parameters unchanged. This design results in a highly efficient training procedure since, although the sequence length is slightly extended by the Robustness Tokens and each batch requires running an adversarial attack algorithm, the total count of parameters to be tuned is much lower than that of the VFM and convergence takes little optimization steps. Since the focus of this work is the adversarial robustness of foundation models, we select task-agnostic attacks to maximally robustify the features extracted by the foundation model and avoid making any assumption on the particular downstream task the model will be used for. However, the proposed method can easily be applied to task-specific adversarial attacks by simply swapping the adversarial attack $A(f, \mathbf{x})$ with a task-specific attack $A(f, \mathbf{x}, \mathbf{y})$, where $\mathbf{y}$ is the appropriate label for the downstream task for sample $\mathbf{x}$. We schematically show our methodology in \cref{fig:training} and provide a Pytorch-like pseudocode in \cref{fig:pseudocode}.

Since we only modify the input to the model rather than its parameters, Robustness Tokens are effectively a novel defense strategy orthogonal to classical adversarial training methodologies. This means that Robustness Tokens can be used in combination with existing defense strategies.

\section{Results}
\label{sec:results}

\subsection{Experimental Setup}
\label{sec:setup}
We train Robustness Tokens only on a small fraction (1600 images) of the ImageNet \cite{imagenet} dataset for all 8 available pre-trained DiNOv2 \cite{dinov2} backbone models of four sizes (small, base, large, giant) both with and without register tokens \cite{registers}. Only the output patch and class tokens are considered features, whereas the output register tokens and Robustness Tokens are not and thus are not used in the computation of the loss as shown in \cref{fig:training}. We train Robustness Tokens using the Adam optimizer, with the learning rate set to $0.001$, $\beta_1=0.9$ and $\beta_2=0.999$. Training converges in a few steps despite a low batch size of 8. For all models, we train 10 Robustness Tokens. We conduct an ablation on the number of tokens in \cref{sec:token_ablation}.

We focus our studies on the DiNOv2 backbone model because a) It is the state-of-the-art SSL vision model when evaluated on Imagenet linear classification and among the most popular VFMs to date and b) we interestingly find that DiNOv2 is extremely vulnerable even to task-agnostic adversarial attacks. These two facts combined make the study of adversarial training for DiNOv2 a very urgent one, so we decided to focus and conduct extensive studies on DiNOv2, leaving a more extensive study of the efficacy of Robustness Tokens for other models for future work.

\subsubsection{Adversarial attacks}
In our experiments, we focus on task-agnostic adversarial attacks that maximally disrupt the feature representation of the foundation model, which allows us to make the foundation model more robust across multiple downstream tasks at once. Notice that our method is not limited to these attacks, and it might be used to robustify foundation models against more specific kinds of attacks. In particular, we obtain the adversarial samples using white-box PGD attacks \cite{pgd} and minimizing the following objective:

\begin{equation}
    \loss_{\text{attack}}(\xadv) = \frac{f(\x) \cdot f(\xadv)}{||f(\x)|| \ ||f(\xadv)||} - ||\x - \xadv||_2^2
\end{equation}

such that the attack maximally perturbs the features extracted by the model according to the cosine similarity while minimally increasing the mean-squared error between the original and adversarial images. We run all attacks for 30 steps setting $\epsilon_{\infty} = \frac{8}{255}$ and ensure that the PSNR between clean and adversarial samples is never below 40.0dB. We also quantize the adversarial attacks to make sure that the obtained tensors are valid png pictures. Our code is publicly available at \url{https://github.com/BrianPulfer/robustness-tokens}.

\subsubsection{Hardware}
We run our experiments on a CentOS SLURM cluster featuring various nodes. We use NVIDIA A100 80GB and RTX 3080 TI 24GB GPUs.

\subsection{Downstream Performances}
To guarantee that the addition of new tokens does not result in an important degradation of downstream performances, we compare the classification and segmentation capabilities of DiNOv2 models that make use of our newly trained Robustness Tokens against the original versions. To do so, we train linear heads on the ImageNet-1k and ADE20K datasets starting from the original DiNOv2 backbone checkpoints and our modified versions using Robustness Tokens with the original DiNOv2 codebase. We fine-tune a total of 16 heads per task, which include all combinations of 4 model sizes, 2 options for whether to use register tokens as in \cite{registers}, and 2 options for whether to use our Robustness Tokens and report results in \cref{tab:downstream}.

\begin{table}[ht]
\centering
\caption{Performances on downstream tasks. For both classification and segmentation tasks, we train linear heads on top of the extracted features. We evaluate classification and segmentation performances on ImageNet \cite{imagenet} and ADE20K \cite{ade20k} datasets and report accuracy and mIoU respectively.}
\begin{tabular}{ccc}
\textbf{Model} & \multicolumn{2}{c}{\textbf{Performance}} \\ \hline
               & \textit{Classification}  & \textit{Segmentation} \\ \hline
DiNOv2-S & 80.0 & 41.0 \\
DiNOv2-B & 83.4 & 45.1 \\
DiNOv2-L & 85.5 & 45.1 \\
DiNOv2-G & 85.2 & 46.6 \\
DiNOv2-S + reg & 79.8 & 40.4 \\
DiNOv2-B + reg & 83.7 & 45.8 \\
DiNOv2-L + reg & 86.1 & 46.6 \\
DiNOv2-G + reg & 86.3 & 46.8 \\
\hline
DiNOv2-S + rob \textbf{(ours)} & 78.5 & 40.6 \\
DiNOv2-B + rob \textbf{(ours)} & 83.1 & 45.0 \\
DiNOv2-L + rob \textbf{(ours)} & 84.2 & 45.5 \\
DiNOv2-G + rob \textbf{(ours)} & 85.6 & 47.2 \\
DiNOv2-S + reg + rob \textbf{(ours)} & 79.2 & 40.9 \\
DiNOv2-B + reg + rob \textbf{(ours)} & 83.1 & 45.8 \\
DiNOv2-L + reg + rob \textbf{(ours)} & 85.9 & 46.7 \\
DiNOv2-G + reg + rob \textbf{(ours)} & 86.1 & 47.5 \\
\hline
\end{tabular}
\label{tab:downstream}
\end{table}

We notice that the downstream performances are comparable despite adding Robustness Tokens after the pre-training stage of DiNOv2. In particular, while classification accuracy slightly deteriorates on average ($-0.5$), the mean interception over union is actually marginally improved ($+0.2$). This finding is insightful, as it shows that it is possible to append additional tokens that do not represent an image to a Vision Transformer and exceed its usual context size without major consequences on the obtained results. We believe that this finding can potentially enable research on how the input sequences fed to Vision Transformer models can be modified to obtain certain desired behaviors beyond adversarial robustness.

\subsection{Robustness to Attacks}
We further study how performances are affected in the face of adversarial attacks for the same datasets. Since we assume Robustness Tokens to be kept secret, adversarial attacks are crafted without taking into consideration the concatenation with Robustness Tokens. However, since we only modify the input to the transformer keeping the architecture and parameters unchanged and assume the attacker has full knowledge of the model and training data, we define all attacks as white-box.

We examine attacks that maximally perturb feature space representation, as done during training, as well as classical classification and semantic segmentation attacks. For feature space corruptions, we measure the average cosine similarity between the original features and those extracted from the adversarial images crafted using adversarial attacks in feature space on the ImageNet dataset. For classification and semantic segmentation, we measure accuracy and mIoU respectively on images attacked adversarially using PGD aiming to maximize cross-entropy loss for the ImageNet and ADE20K datasets respectively. We report the results in \cref{tab:robustness}.

\begin{table}[ht]
\centering
\caption{Robustness to adversarial attacks in feature space, classification, and semantic segmentation.}
\begin{tabular}{cccc}
\textbf{Model} & \multicolumn{3}{c}{\textbf{Robustness}} \\ \hline
               & \textit{Features}  & \textit{Classification}  & \textit{Segmentation} \\ \hline
DiNOv2-S & 0.09 & 0.0 &  2.8 \\
DiNOv2-B & 0.05 & 0.0 & 3.1 \\
DiNOv2-L & 0.06 & 0.3 & 4.5 \\
DiNOv2-G & 0.12 & 0.3 & 4.7 \\
DiNOv2-S + reg & 0.01 & 0.0 & 2.1 \\
DiNOv2-B + reg & 0.03 & 0.1 & 3.0 \\
DiNOv2-L + reg & 0.03 & 0.6 & 4.6 \\
DiNOv2-G + reg & 0.08 & 0.9 & 4.2 \\
\hline
DiNOv2-S + rob \textbf{(ours)} & 0.93 & 31.9 & 24.6 \\
DiNOv2-B + rob \textbf{(ours)} & 0.92 & 50.0 & 23.4 \\
DiNOv2-L + rob \textbf{(ours)} & 0.89 & 62.9 & 21.2 \\
DiNOv2-G + rob \textbf{(ours)} & 0.89 & 63.1 & 23.3 \\
DiNOv2-S + reg + rob \textbf{(ours)} & 0.93 & 30.5 & 22.7 \\
DiNOv2-B + reg + rob \textbf{(ours)} & 0.92 & 49.7 & 25.9 \\
DiNOv2-L + reg + rob \textbf{(ours)} & 0.83 & 58.7 & 16.2 \\
DiNOv2-G + reg + rob \textbf{(ours)} & 0.90 & 69.9 & 25.7 \\
\hline
\end{tabular}
\label{tab:robustness}
\end{table}

From \cref{tab:robustness} we observe that features extracted with DiNOv2 can be easily put close to orthogonal directions with a task-agnostic white-box adversarial attack, resulting in a total degradation of performances for both classification and segmentation tasks. However, after adding the trained Robustness Tokens, the cosine similarity with respect to the original extracted features is much higher, mitigating most of the adversarial effect. We also observe that downstream performance facing white-box adversarial attacks is greatly enhanced for both classification and segmentation.

Interestingly, we notice that bigger models benefit the most from the use of Robustness Tokens for image classification. Our intuition is that adversarial attacks take advantage of massive activations \cite{sun2024massive} recently observed in larger models, and that Robustness Tokens restore such activations to usual values. We study this phenomenon in \cref{sec:massive}.

\begin{figure}[ht]
    \centering
    \includegraphics[width=\linewidth]{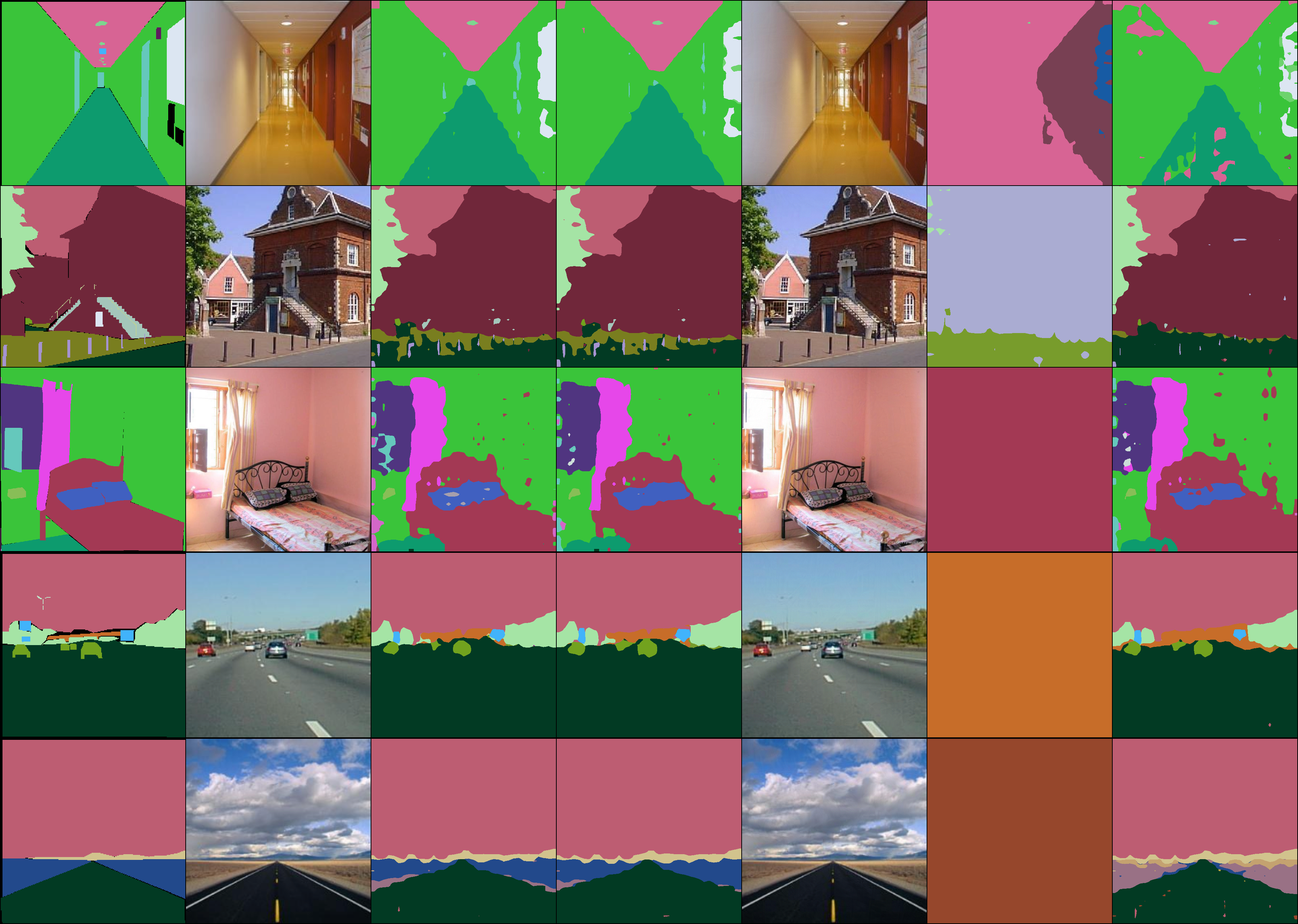}
    \caption{Segmentation predictions of vanilla and enhanced DiNOv2 ViT-B/14 on clean and adversarial samples from the ADE20K dataset. From left to right: Ground truth segmentation mask, clean sample, prediction on the clean sample, prediction with Robustness Tokens on the clean sample, adversarial sample, prediction on the adversarial sample, prediction with Robustness Tokens on the adversarial sample.}
    \label{fig:segmentation}
\end{figure}

In \cref{fig:segmentation}, we show the segmentation capabilities of DiNOv2 against clean and adversarial samples with and without the use of Robustness Tokens. While segmentation masks on clean samples are similar for both the original and robustified models, those obtained from adversarial samples are completely perturbed for original models, whereas robustified models still retain relatively high-quality segmentation masks.

\subsection{Ablation on the Number of Tokens}
\label{sec:token_ablation}
\begin{figure}
\centering
\begin{subfigure}{.5\textwidth}
  \centering
  \includegraphics[width=\textwidth]{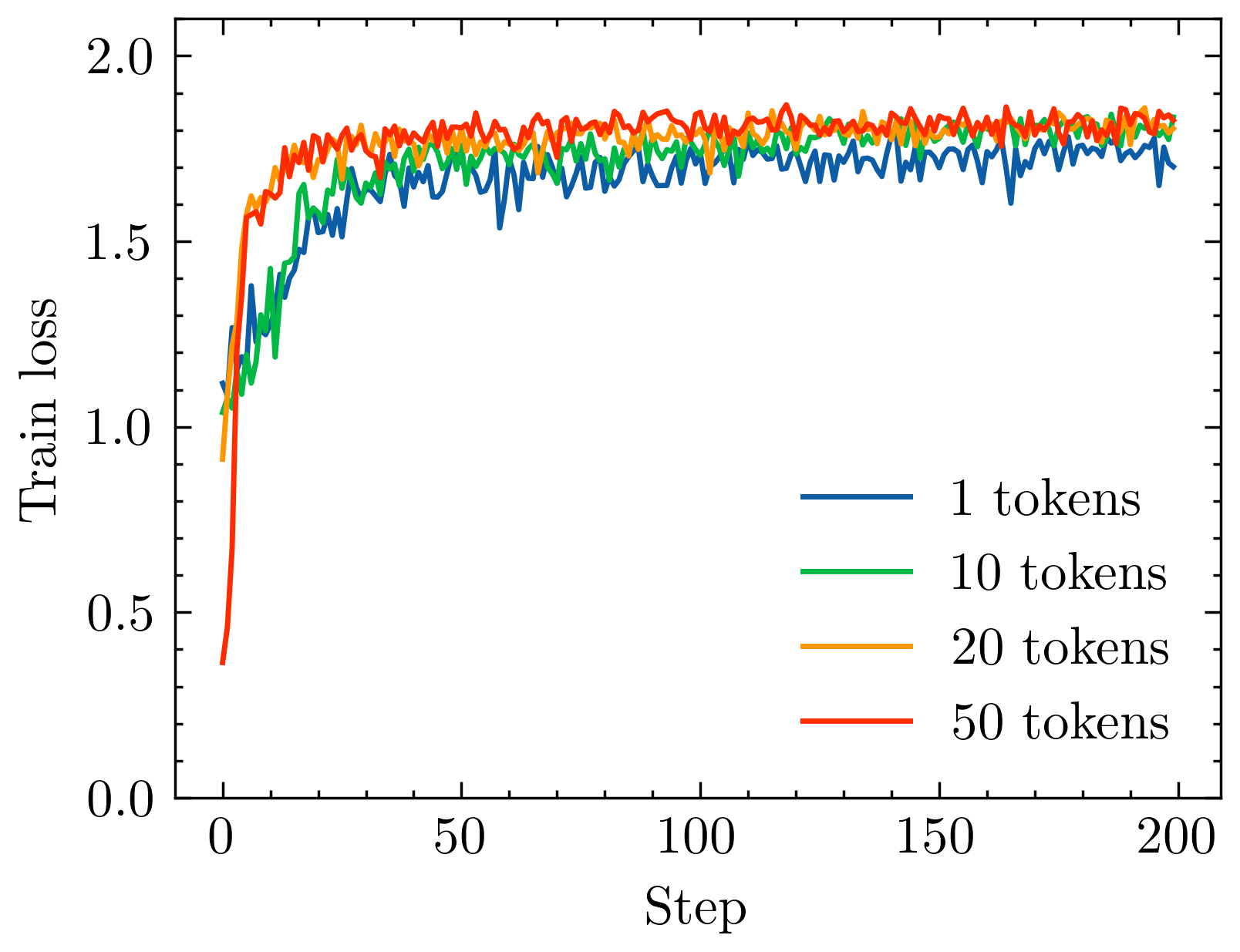}
  \caption{ViT-S/14}
  \label{fig:sub1}
\end{subfigure}%
\begin{subfigure}{.5\textwidth}
  \centering
  \includegraphics[width=\textwidth]{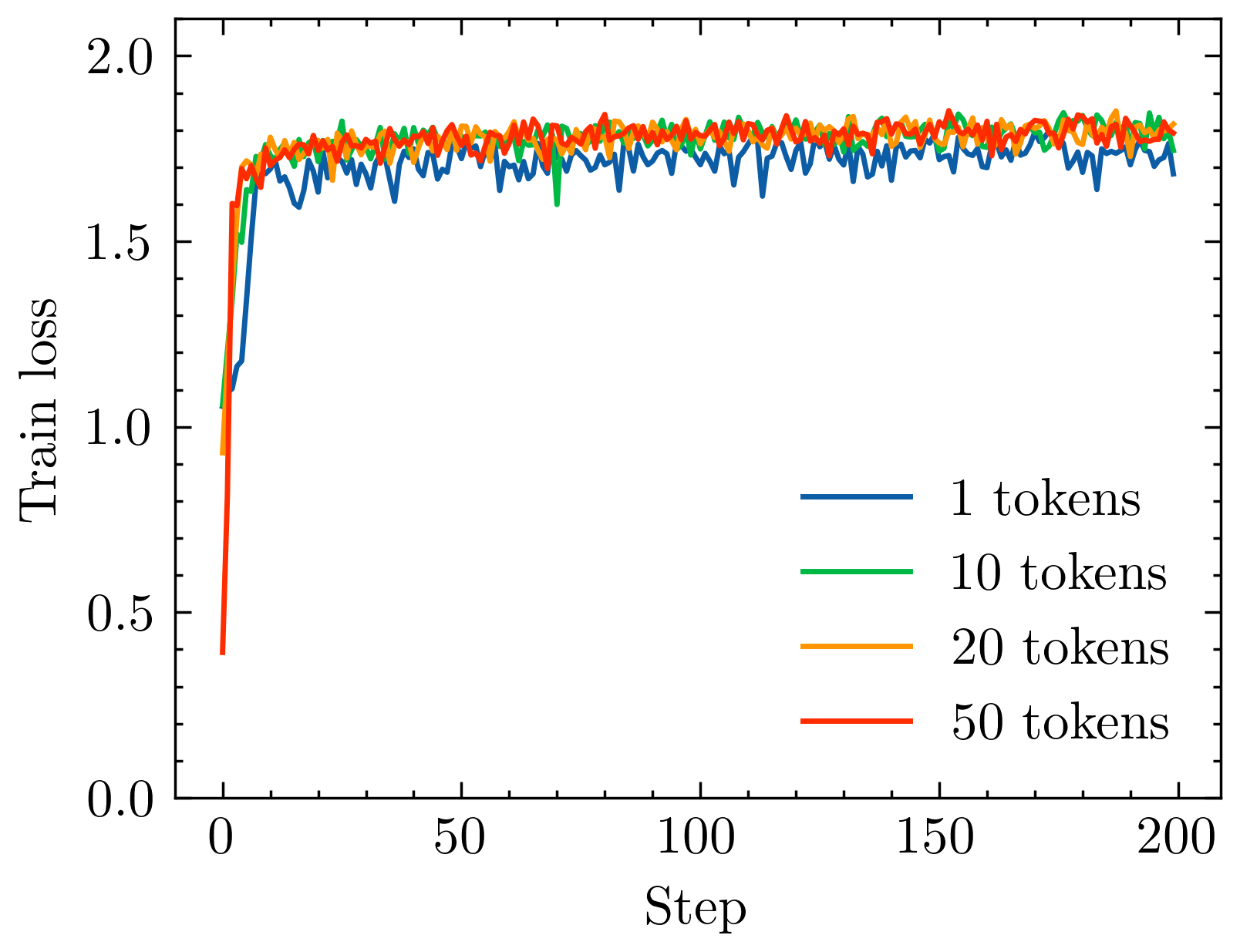}
  \caption{ViT-B/14}
  \label{fig:sub2}
\end{subfigure}
\begin{subfigure}{.5\textwidth}
  \centering
  \includegraphics[width=\textwidth]{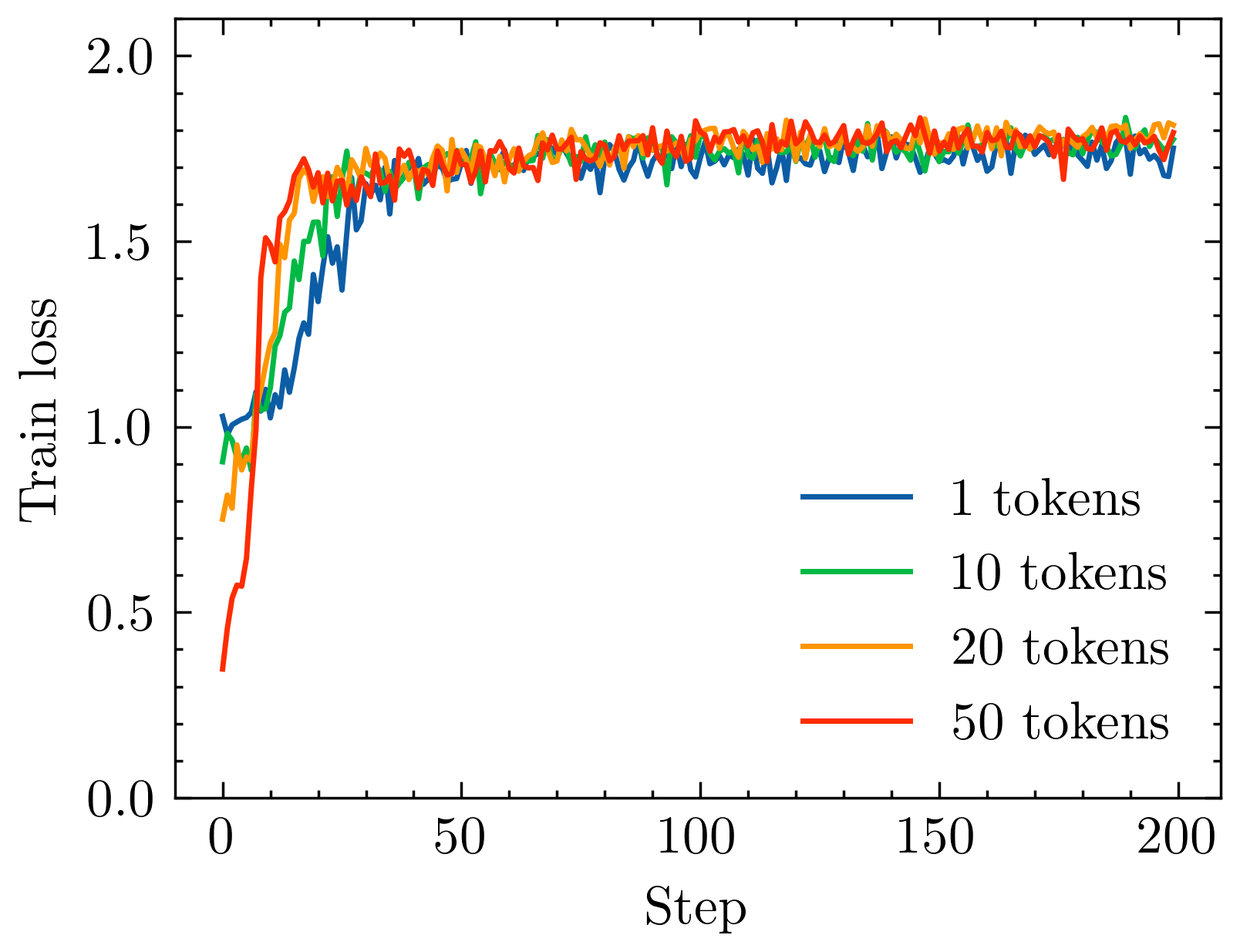}
  \caption{ViT-L/14}
  \label{fig:sub3}
\end{subfigure}%
\begin{subfigure}{.5\textwidth}
  \centering
  \includegraphics[width=\textwidth]{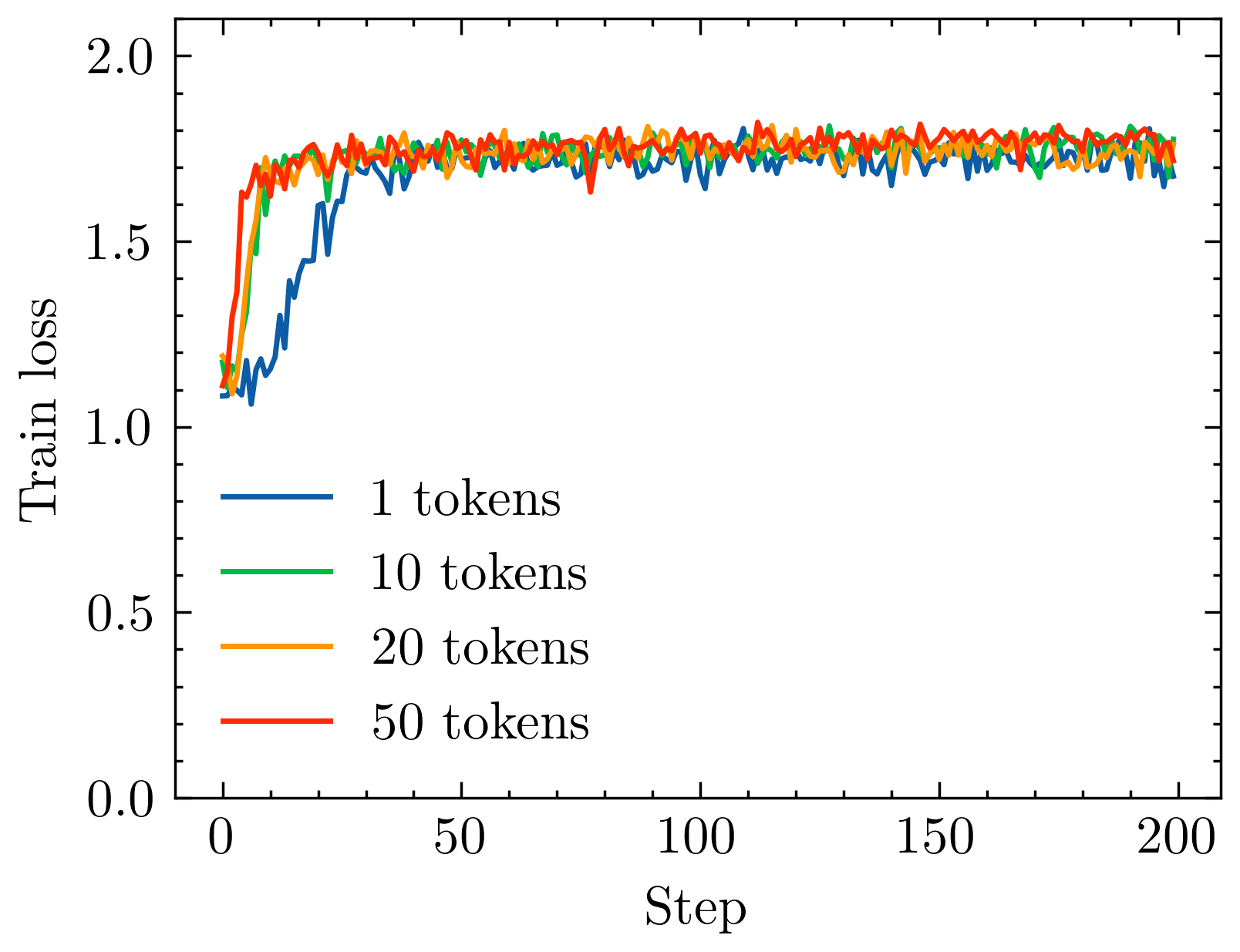}
  \caption{ViT-G/14}
  \label{fig:sub4}
\end{subfigure}
\caption{Training curves for training of Robustness Tokens for DiNOv2. We train small, base, large, and giant models with 1, 10, 20, and 50 Robustness Tokens. Training converges within a few steps in all cases.}
\label{fig:token_ablation}
\end{figure}

We conduct an ablation study on the amount of register tokens needed to robustify DiNOv2. We train all four original DiNOv2 models with 1, 10, 20, and 50 Robustness Tokens and show the training loss through steps in \cref{fig:token_ablation}.

From this ablation, we notice that even a single robustness token can greatly correct the orientation of features extracted with DiNOv2. This result parallels that of \cite{registers}, who found that even one register token could correct for artifacts in attention masks. We also note that the use of a few tokens yields better training loss, but that little to no gain seems to be obtained with more than 10 tokens for all model sizes.

\subsection{Training Efficiency}
Through \cref{fig:token_ablation} we also highlight the high efficiency of training Robustness Tokens. We observe that Robustness Tokens converge well before the first epoch on the ImageNet dataset is over. In fact, training saturates in just under 200 steps of training on Imagenet with a batch size of 8 for all models and the number of Robustness Tokens used.

Even though obtaining adversarial samples is a computationally intense task that involves computing gradients with respect to the input through the whole network for multiple steps (30 in our study), the low parameter count of Robustness Tokens allows them to converge in relatively few total gradient computations ($30 \times 200 + 200 = 6'200$). Furthermore, the computational budget required for training can be further lowered by trading off adversarial samples' effectiveness.

\subsection{Applicability to different models}
We study whether the benefits of Robustness Tokens apply to other models. To do so, similarly to \cite{registers}, we consider OpenCLIP \cite{opencliprepo} and DEIT-III \cite{touvron2022deit} models.

\begin{figure}
\centering
\begin{subfigure}{.5\textwidth}
  \centering
  \includegraphics[width=\textwidth]{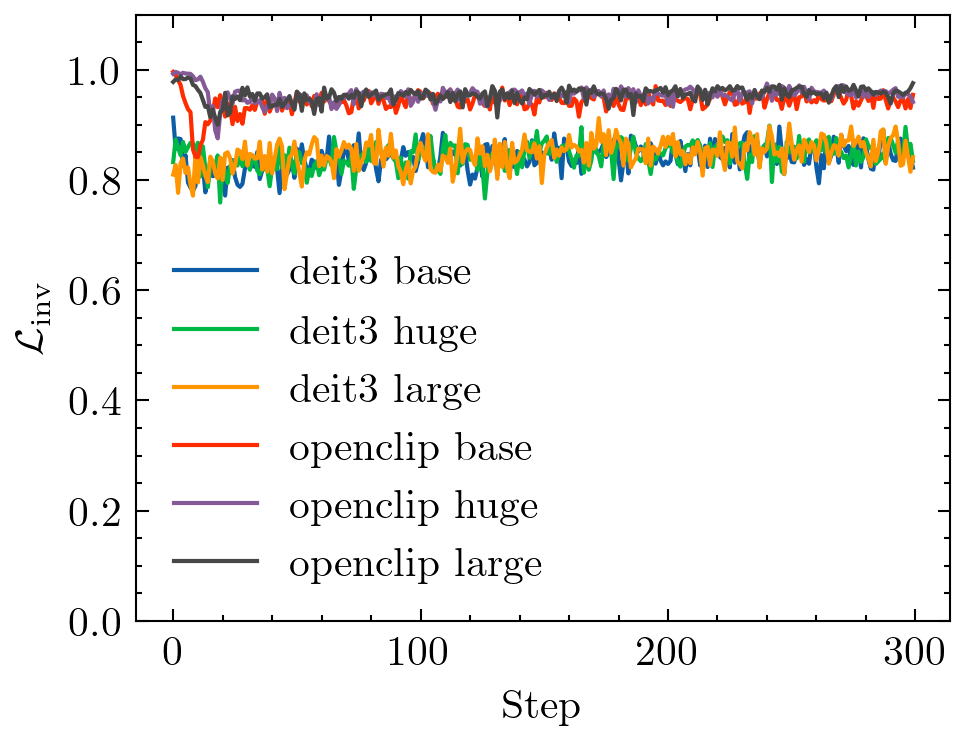}
  \caption{Invariance loss term $\linv$}
  \label{fig:model_abl_sub1}
\end{subfigure}%
\begin{subfigure}{.5\textwidth}
  \centering
  \includegraphics[width=\textwidth]{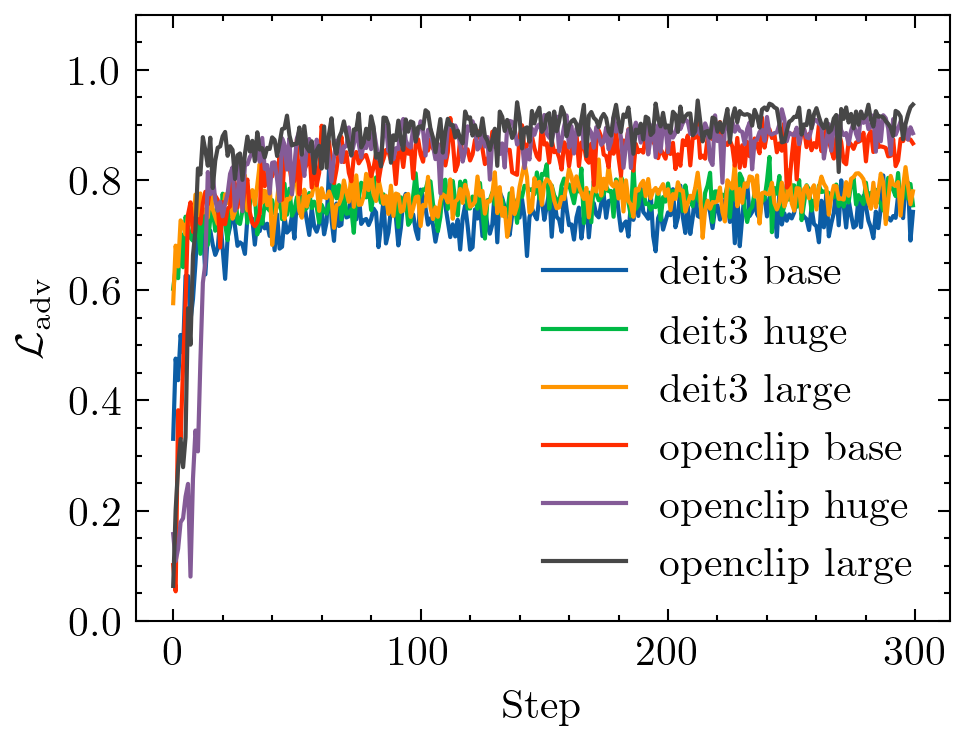}
  \caption{Adversarial loss term $\ladv$}
  \label{fig:model_abl_sub2}
\end{subfigure}
\caption{$\linv$ and $\ladv$ terms through training for base, large and huge DEIT-III and OpenCLIP models. While the $\linv$ term is relatively stable through training, the $\ladv$ is quickly maximized.}
\label{fig:model_ablation}
\end{figure}

As done for DiNOv2, we train 10 Robustness Tokens for the base, large, and huge OpenCLIP and DEIT-III models. In \cref{fig:model_ablation}, we observe that Robustness Tokens greatly increase the cosine similarity of features extracted from a clean and adversarial sample for both OpenCLIP and DEIT-III models. As for DiNOv2, the $\ladv$ loss term is quickly maximized, whereas the $\linv$ term remains relatively stable throughout training. 

\begin{table}
    \centering
    \caption{Average cosine similarity between clean and adversarial samples for DEIT-III and OpenCLIP models with and without Robustness Tokens. The average is taken over the last 100 batches used during training with batch size 8.}
    \begin{tabular}{c|c|c}
         \textbf{Model} & \textbf{Regular} & \textbf{Robustified} \\
         \hline
         DEIT-III Base & 0.16 ± 0.04 & \textbf{0.74} ± 0.03 \\
         DEIT-III Large & 0.22 ± 0.03 & \textbf{0.78} ± 0.02 \\
         DEIT-III Huge & 0.23 ± 0.03 & \textbf{0.77} ± 0.02 \\
         OpenCLIP Base & -0.02 ± 0.05 & \textbf{0.86} ± 0.02 \\
         OpenCLIP Large & 0.13 ± 0.06 & \textbf{0.91} ± 0.02 \\
         OpenCLIP Huge & 0.10 ± 0.07 &  \textbf{0.89} ± 0.02 \\
         \hline
    \end{tabular}
    \label{tab:model_ablation}
\end{table}

In \cref{tab:model_ablation}, we further report the average cosine similarity, over the last 100 batches used in training, between clean and adversarial samples for the regular pre-trained foundation models and those robustified with Robustness Tokens.

Interestingly, we observe that Robustness Tokens are consistently more effective on OpenCLIP than DEIT-III since, in terms of cosine similarity, original OpenCLIP models seem more fragile to adversarial attacks than DEIT-III models, but end up becoming more robust after the introduction of Robustness Tokens. This suggests that different models might benefit differently from the introduction of Robustness Tokens.

\subsection{Generalization to multiple attacks}
Although it would be possible to robustify models against a multitude of attacks rather than just one, computing highly effective adversarial samples is typically a computationally intensive task. The computational cost of adversarial training with Robustness Tokens may not need to increase if such tokens could generalize across different adversarial attack families.

We test whether the Robustness Tokens trained to defend against PGD attacks generalize to different adversarial attacks by measuring classification accuracy with respect to adversarial attacks crafted using AutoAttack \cite{autoattack, kim2020torchattacks}, the attack methodology used by RobustBench \cite{croce2020robustbench}, on the original DiNOv2 models in \cref{tab:attacks_ablation}.

\begin{table}[ht]
\centering
\caption{Classification accuracy on the ImageNet validation set with clean and adversarial samples obtained with AutoAttack. Our method generalizes to different kinds of adversarial attacks than those used for training the Robustness Tokens.}
\begin{tabular}{ccc}
\textbf{Model} & \textbf{Regular}  & \textbf{Robustified} \\ \hline
DiNOv2-S & 0.5 & \textbf{29.5}\\
DiNOv2-B & 0.9 & \textbf{46.4} \\
DiNOv2-L & 1.8 & \textbf{60.1} \\
DiNOv2-S-reg & 0.3 & \textbf{24.2}\\
DiNOv2-B-reg & 0.8 & \textbf{43.0} \\
DiNOv2-L-reg & 1.9 & \textbf{59.6}\\
\hline
\end{tabular}
\label{tab:attacks_ablation}
\end{table}

From \cref{tab:attacks_ablation}, we observe that Robustness Tokens do indeed generalize to different adversarial attacks since AutoAttack uses an ensemble of parameter-free attacks. Furthermore, the trend of larger models benefitting the most from Robustness Tokens is once again observed.

\subsection{Massive Activations and Robustness Tokens}
\label{sec:massive}
We hypothesize that adversarial attacks might in part take advantage of the recently observed massive activations in transformers \cite{sun2024massive}.

\begin{figure}
\centering
\begin{subfigure}{.33\linewidth}
  \centering
  \includegraphics[width=\textwidth]{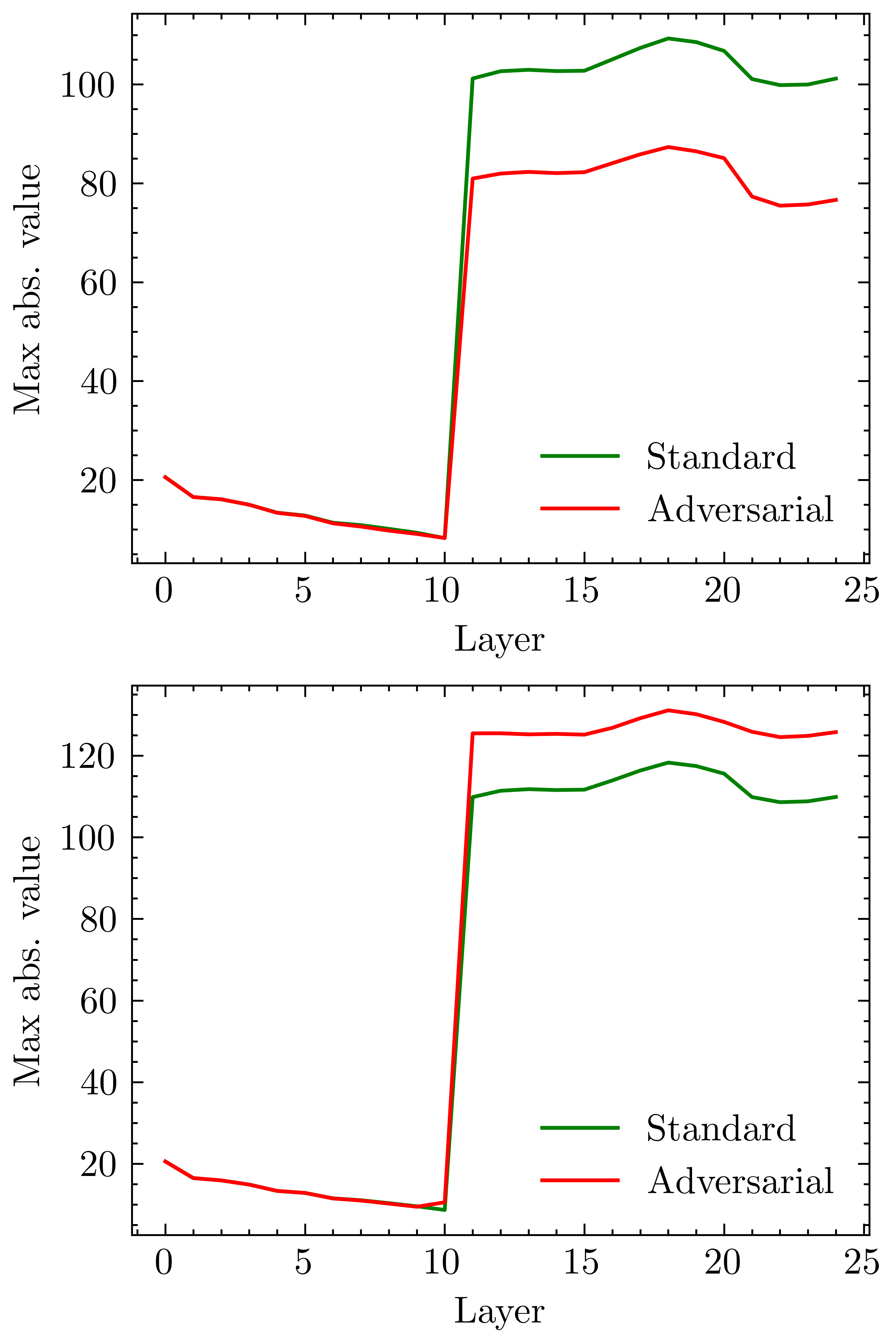}
  \caption{OpenCLIP}
  \label{fig:maclip}
\end{subfigure}%
\begin{subfigure}{.33\linewidth}
  \centering
  \includegraphics[width=\textwidth]{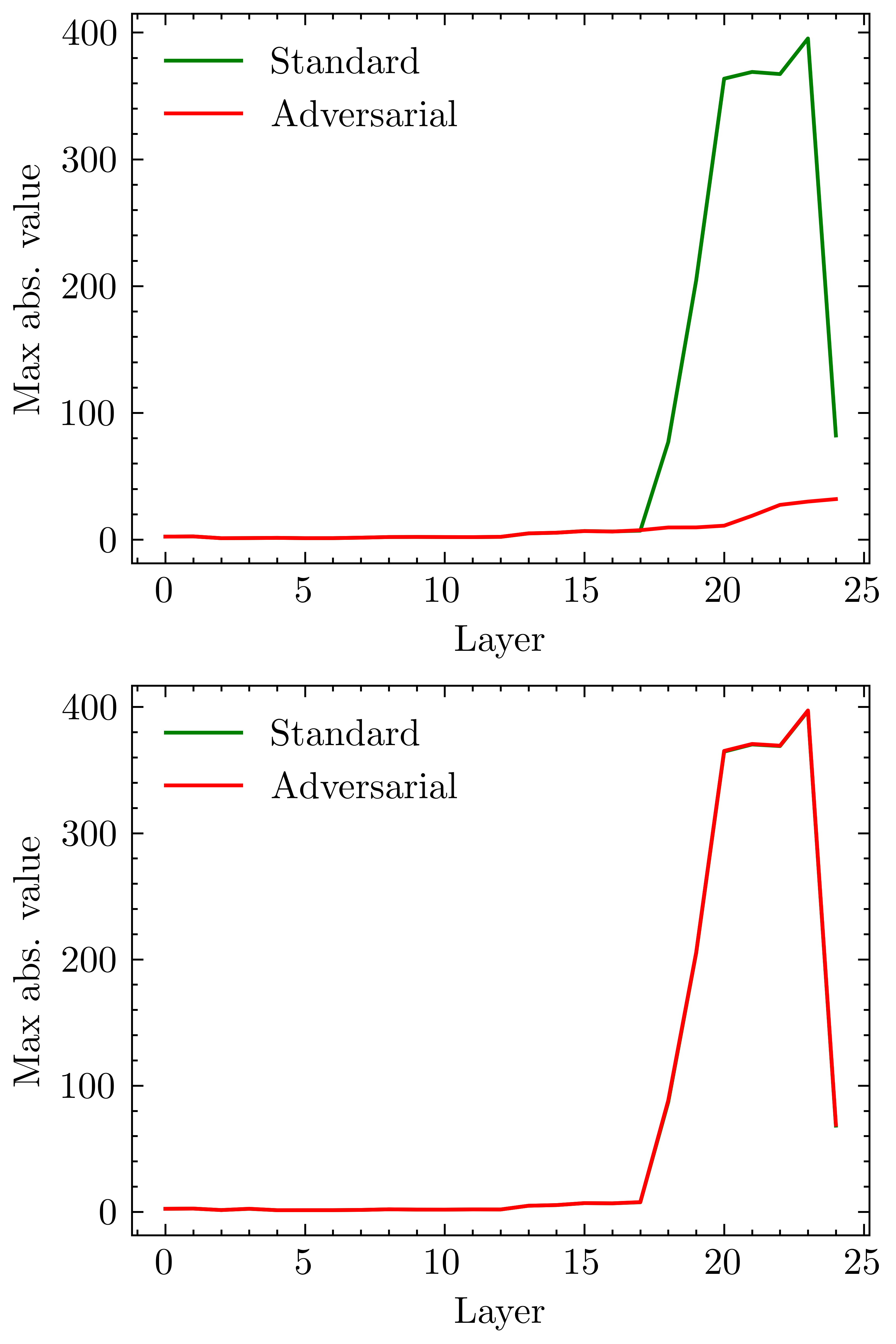}
  \caption{DiNOv2}
  \label{fig:madino}
\end{subfigure}%
\begin{subfigure}{.33\linewidth}
  \centering
  \includegraphics[width=\textwidth]{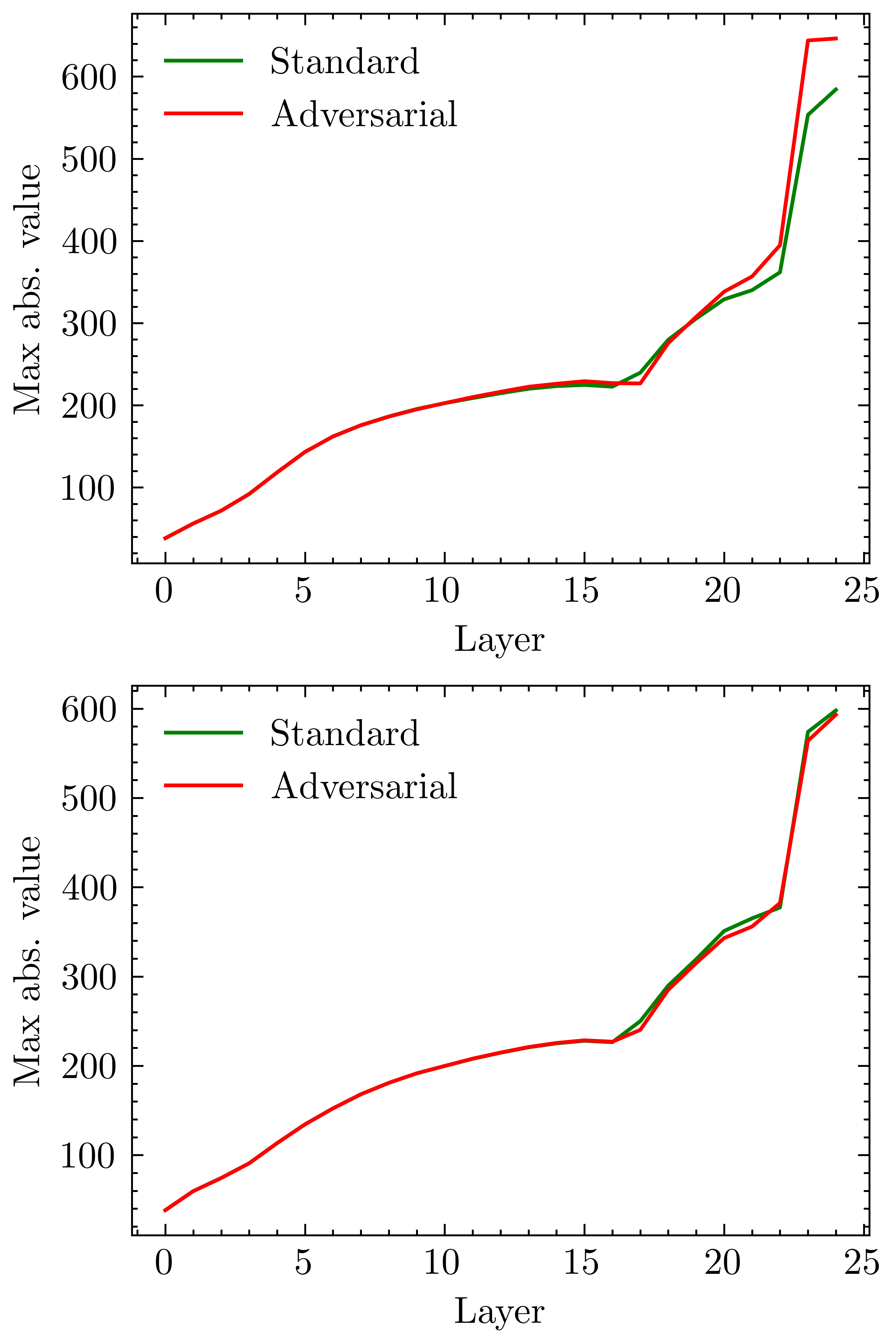}
  \caption{DEIT-III}
  \label{fig:madeit}
\end{subfigure}
\caption{Massive activations before (\textit{first row}) and after, (\textit{second row}), the introduction of robustness tokens for OpenCLIP, DiNOv2 and DEIT-III ViT large models. For each layer, we report the maximum absolute value.}
\label{fig:massive_acts}
\end{figure}

To assess the impact of adversarial attacks and robustness tokens to massive activations, in \cref{fig:massive_acts}, for each layer in the ViT-L, we show the maximum absolute value present in class or patch tokens for a single sample and its adversary counterpart for  OpenCLIP, DiNOv2, and DEIT-III models.

In general, from \cref{fig:massive_acts}, we confirm that adversarial attacks seem to take advantage of massive activations observed in the larger models. We also observe that robustness tokens bridge the gap between massive activations obtained with clean samples and those obtained with adversarial samples for all models.

We report that for OpenCLIP and DiNOv2 models, adversarial attacks manage to set huge activations to much lower values. This is in line with \cite{sun2024massive}, who observed that when massive activations were manually set to zero, the performances of the models degraded consistently. The introduction of Robustness Tokens brings massive activations for adversarial samples to nominal values for DiNOv2, whereas for OpenCLIP such activations seem to be even increased.

For DEIT-III, once again, we observe that adversarial attacks manage to modify the maximum massive activation. Unlike DiNOv2 and OpenCLIP however, adversarial attacks for DEIT-III seem to increase the magnitude of the maximum massive activation rather than decrease it. Furthermore, the gap between massive activations obtained with clean and adversarial samples is much thinner, and it is mostly marked in the very last few layers hinting that perhaps adversarial attacks could not exploit massive activations as thoroughly as for DiNOv2 and OpenCLIP, and that robustness tokens still mitigate the adversarial effect as they do for smaller models where such massive activations are not observed \cite{sun2024massive}.

\section{Discussion}
\label{sec:discussion}
A possible limitation of our method is that, given how quickly robustness tokens can be trained, an attacker might be able to train such tokens and use them to obtain a more informed and possibly more effective attack. However, many combinations emerge as for how many, with which parameters, and in which layers the defender may use the secret tokens. Further research is needed to better understand the extent to which robustness tokens can be inferred by an attacker. Finally, while our method can, in principle, be used on any transformer and sequence model, further studies of their applicability and effectiveness in other domains are needed, and we leave them for future work.

\section{Conclusions}
\label{sec:conclusions}
In this work, we presented Robustness Tokens (\texttt{ROB} tokens), a novel technique that, instead of modifying model parameters as traditionally done in adversarial training, learns a few additional and secret transformer tokens that keep the representational capabilities of transformer models for both clean and adversarial samples.

In our experiments, we show that Robustness Tokens make ViT models more robust to adversarial attacks crafted without the knowledge of such tokens while keeping downstream performance almost unaffected. We show that Robustness Tokens can be trained at a low computational cost, given the fast convergence of the learning algorithm for so few parameters. We find that Robustness Tokens are applicable to a multitude of ViT models, and that they generalize to different adversarial attacks. We also discover that adversarial attacks take advantage of massive activations \cite{sun2024massive}, and that robustness tokens learn to restore them, while also being effective for smaller models where such activations are not observed.

We hope that our work will inspire research on adversarial robustness on the less explored axis of input data and transformer tokens, which we believe to be a promising research direction beyond adversarial robustness to be investigated.

\bibliographystyle{splncs04}
\bibliography{main}

\begin{thebibliography}{10}
\providecommand{\url}[1]{\texttt{#1}}
\providecommand{\urlprefix}{URL }
\providecommand{\doi}[1]{https://doi.org/#1}

\bibitem{arnab2021vivit}
Arnab, A., Dehghani, M., Heigold, G., Sun, C., Lučić, M., Schmid, C.: Vivit: A video vision transformer (2021)

\bibitem{msn}
Assran, M., Caron, M., Misra, I., Bojanowski, P., Bordes, F., Vincent, P., Joulin, A., Rabbat, M., Ballas, N.: Masked siamese networks for label-efficient learning. In: Avidan, S., Brostow, G., Ciss{\'e}, M., Farinella, G.M., Hassner, T. (eds.) Computer Vision -- ECCV 2022. pp. 456--473. Springer Nature Switzerland, Cham (2022)

\bibitem{ijepa}
Assran, M., Duval, Q., Misra, I., Bojanowski, P., Vincent, P., Rabbat, M., LeCun, Y., Ballas, N.: Self-supervised learning from images with a joint-embedding predictive architecture. In: Proceedings of the IEEE/CVF Conference on Computer Vision and Pattern Recognition (CVPR). pp. 15619--15629 (June 2023)

\bibitem{awais2023foundational}
Awais, M., Naseer, M., Khan, S., Anwer, R.M., Cholakkal, H., Shah, M., Yang, M.H., Khan, F.S.: Foundational models defining a new era in vision: A survey and outlook. arXiv preprint arXiv:2307.13721  (2023)

\bibitem{bahri2021explaining}
Bahri, Y., Dyer, E., Kaplan, J., Lee, J., Sharma, U.: Explaining neural scaling laws. arXiv preprint arXiv:2102.06701  (2021)

\bibitem{bai2021recent}
Bai, T., Luo, J., Zhao, J., Wen, B., Wang, Q.: Recent advances in adversarial training for adversarial robustness. arXiv preprint arXiv:2102.01356  (2021)

\bibitem{balestriero2023cookbook}
Balestriero, R., Ibrahim, M., Sobal, V., Morcos, A., Shekhar, S., Goldstein, T., Bordes, F., Bardes, A., Mialon, G., Tian, Y., et~al.: A cookbook of self-supervised learning. arXiv preprint arXiv:2304.12210  (2023)

\bibitem{ban2022pre}
Ban, Y., Dong, Y.: Pre-trained adversarial perturbations. Advances in Neural Information Processing Systems  \textbf{35},  1196--1209 (2022)

\bibitem{bommasani2021opportunities}
Bommasani, R., Hudson, D.A., Adeli, E., Altman, R., Arora, S., von Arx, S., Bernstein, M.S., Bohg, J., Bosselut, A., Brunskill, E., et~al.: On the opportunities and risks of foundation models. arXiv preprint arXiv:2108.07258  (2021)

\bibitem{carlini2017towards}
Carlini, N., Wagner, D.: Towards evaluating the robustness of neural networks. In: 2017 ieee symposium on security and privacy (sp). pp. 39--57. Ieee (2017)

\bibitem{cae}
Chen, X., Ding, M., Wang, X., Xin, Y., Mo, S., Wang, Y., Han, S., Luo, P., Zeng, G., Wang, J.: Context autoencoder for self-supervised representation learning (2022)

\bibitem{croce2020robustbench}
Croce, F., Andriushchenko, M., Sehwag, V., Debenedetti, E., Flammarion, N., Chiang, M., Mittal, P., Hein, M.: Robustbench: a standardized adversarial robustness benchmark. arXiv preprint arXiv:2010.09670  (2020)

\bibitem{autoattack}
Croce, F., Hein, M.: Reliable evaluation of adversarial robustness with an ensemble of diverse parameter-free attacks. In: International conference on machine learning. pp. 2206--2216. PMLR (2020)

\bibitem{registers}
Darcet, T., Oquab, M., Mairal, J., Bojanowski, P.: Vision transformers need registers (2023)

\bibitem{demontis2019adversarial}
Demontis, A., Melis, M., Pintor, M., Jagielski, M., Biggio, B., Oprea, A., Nita-Rotaru, C., Roli, F.: Why do adversarial attacks transfer? explaining transferability of evasion and poisoning attacks. In: 28th USENIX security symposium (USENIX security 19). pp. 321--338 (2019)

\bibitem{imagenet}
Deng, J., Dong, W., Socher, R., Li, L.J., Li, K., Fei-Fei, L.: Imagenet: A large-scale hierarchical image database. In: 2009 IEEE conference on computer vision and pattern recognition. pp. 248--255. Ieee (2009)

\bibitem{qlora}
Dettmers, T., Pagnoni, A., Holtzman, A., Zettlemoyer, L.: Qlora: Efficient finetuning of quantized llms. Advances in Neural Information Processing Systems  \textbf{36} (2024)

\bibitem{devlin2018bert}
Devlin, J., Chang, M.W., Lee, K., Toutanova, K.: Bert: Pre-training of deep bidirectional transformers for language understanding. arXiv preprint arXiv:1810.04805  (2018)

\bibitem{vit}
Dosovitskiy, A., Beyer, L., Kolesnikov, A., Weissenborn, D., Zhai, X., Unterthiner, T., Dehghani, M., Minderer, M., Heigold, G., Gelly, S., Uszkoreit, J., Houlsby, N.: An image is worth 16x16 words: Transformers for image recognition at scale (2021)

\bibitem{aim}
El-Nouby, A., Klein, M., Zhai, S., Bautista, M.A., Toshev, A., Shankar, V., Susskind, J.M., Joulin, A.: Scalable pre-training of large autoregressive image models (2024)

\bibitem{Fort2021CLIPadversarial}
Fort, S.: Adversarial examples for the openai clip in its zero-shot classification regime and their semantic generalization (Jan 2021), \url{https://stanislavfort.github.io/2021/01/12/OpenAI_CLIP_adversarial_examples.html}

\bibitem{goodfellow2014explaining}
Goodfellow, I.J., Shlens, J., Szegedy, C.: Explaining and harnessing adversarial examples. arXiv preprint arXiv:1412.6572  (2014)

\bibitem{gu2023mamba}
Gu, A., Dao, T.: Mamba: Linear-time sequence modeling with selective state spaces (2023)

\bibitem{hatamizadeh2024vir}
Hatamizadeh, A., Ranzinger, M., Lan, S., Alvarez, J.M., Fidler, S., Kautz, J.: Vir: Towards efficient vision retention backbones (2024)

\bibitem{he2021towards}
He, J., Zhou, C., Ma, X., Berg-Kirkpatrick, T., Neubig, G.: Towards a unified view of parameter-efficient transfer learning. arXiv preprint arXiv:2110.04366  (2021)

\bibitem{mae}
He, K., Chen, X., Xie, S., Li, Y., Doll\'ar, P., Girshick, R.: Masked autoencoders are scalable vision learners. In: Proceedings of the IEEE/CVF Conference on Computer Vision and Pattern Recognition (CVPR). pp. 16000--16009 (June 2022)

\bibitem{NEURIPS2019_a2b15837}
Hendrycks, D., Mazeika, M., Kadavath, S., Song, D.: Using self-supervised learning can improve model robustness and uncertainty. In: Wallach, H., Larochelle, H., Beygelzimer, A., d\textquotesingle Alch\'{e}-Buc, F., Fox, E., Garnett, R. (eds.) Advances in Neural Information Processing Systems. vol.~32. Curran Associates, Inc. (2019), \url{https://proceedings.neurips.cc/paper_files/paper/2019/file/a2b15837edac15df90721968986f7f8e-Paper.pdf}

\bibitem{hestness2017deep}
Hestness, J., Narang, S., Ardalani, N., Diamos, G., Jun, H., Kianinejad, H., Patwary, M.M.A., Yang, Y., Zhou, Y.: Deep learning scaling is predictable, empirically. arXiv preprint arXiv:1712.00409  (2017)

\bibitem{hochreiter1997long}
Hochreiter, S., Schmidhuber, J.: Long short-term memory. Neural computation  \textbf{9}(8),  1735--1780 (1997)

\bibitem{lora}
Hu, E.J., Shen, Y., Wallis, P., Allen-Zhu, Z., Li, Y., Wang, S., Wang, L., Chen, W.: Lora: Low-rank adaptation of large language models. arXiv preprint arXiv:2106.09685  (2021)

\bibitem{opencliprepo}
Ilharco, G., Wortsman, M., Wightman, R., Gordon, C., Carlini, N., Taori, R., Dave, A., Shankar, V., Namkoong, H., Miller, J., Hajishirzi, H., Farhadi, A., Schmidt, L.: Openclip (Jul 2021). \doi{10.5281/zenodo.5143773}, \url{https://doi.org/10.5281/zenodo.5143773}

\bibitem{inkawhich2023adversarial}
Inkawhich, N., McDonald, G., Luley, R.: Adversarial attacks on foundational vision models (2023)

\bibitem{align}
Jia, C., Yang, Y., Xia, Y., Chen, Y.T., Parekh, Z., Pham, H., Le, Q., Sung, Y.H., Li, Z., Duerig, T.: Scaling up visual and vision-language representation learning with noisy text supervision. In: International conference on machine learning. pp. 4904--4916. PMLR (2021)

\bibitem{jiang2024supervised}
Jiang, X., Ge, Y., Ge, Y., Yuan, C., Shan, Y.: Supervised fine-tuning in turn improves visual foundation models. arXiv preprint arXiv:2401.10222  (2024)

\bibitem{kim2020torchattacks}
Kim, H.: Torchattacks: A pytorch repository for adversarial attacks. arXiv preprint arXiv:2010.01950  (2020)

\bibitem{blip}
Li, J., Li, D., Xiong, C., Hoi, S.: Blip: Bootstrapping language-image pre-training for unified vision-language understanding and generation. In: International Conference on Machine Learning. pp. 12888--12900. PMLR (2022)

\bibitem{glip}
Li, L.H., Zhang, P., Zhang, H., Yang, J., Li, C., Zhong, Y., Wang, L., Yuan, L., Zhang, L., Hwang, J.N., et~al.: Grounded language-image pre-training. In: Proceedings of the IEEE/CVF Conference on Computer Vision and Pattern Recognition. pp. 10965--10975 (2022)

\bibitem{li2021prefix}
Li, X.L., Liang, P.: Prefix-tuning: Optimizing continuous prompts for generation. arXiv preprint arXiv:2101.00190  (2021)

\bibitem{flip}
Li, Y., Fan, H., Hu, R., Feichtenhofer, C., He, K.: Scaling language-image pre-training via masking. In: Proceedings of the IEEE/CVF Conference on Computer Vision and Pattern Recognition. pp. 23390--23400 (2023)

\bibitem{lian2024less}
Lian, C., Zhou, H.Y., Yu, Y., Wang, L.: Less could be better: Parameter-efficient fine-tuning advances medical vision foundation models. arXiv preprint arXiv:2401.12215  (2024)

\bibitem{liu2021swin}
Liu, Z., Lin, Y., Cao, Y., Hu, H., Wei, Y., Zhang, Z., Lin, S., Guo, B.: Swin transformer: Hierarchical vision transformer using shifted windows. In: Proceedings of the IEEE/CVF international conference on computer vision. pp. 10012--10022 (2021)

\bibitem{pgd}
Madry, A., Makelov, A., Schmidt, L., Tsipras, D., Vladu, A.: Towards deep learning models resistant to adversarial attacks (2019)

\bibitem{mann2020language}
Mann, B., Ryder, N., Subbiah, M., Kaplan, J., Dhariwal, P., Neelakantan, A., Shyam, P., Sastry, G., Askell, A., Agarwal, S., et~al.: Language models are few-shot learners. arXiv preprint arXiv:2005.14165  (2020)

\bibitem{off_by_one}
Miller, E.: Attention is off by one (2023), \url{https://www.evanmiller.org/attention-is-off-by-one.html}

\bibitem{moosavi2016deepfool}
Moosavi-Dezfooli, S.M., Fawzi, A., Frossard, P.: Deepfool: a simple and accurate method to fool deep neural networks. In: Proceedings of the IEEE conference on computer vision and pattern recognition. pp. 2574--2582 (2016)

\bibitem{nguyen2015deep}
Nguyen, A., Yosinski, J., Clune, J.: Deep neural networks are easily fooled: High confidence predictions for unrecognizable images. In: Proceedings of the IEEE conference on computer vision and pattern recognition. pp. 427--436 (2015)

\bibitem{dinov2}
Oquab, M., Darcet, T., Moutakanni, T., Vo, H., Szafraniec, M., Khalidov, V., Fernandez, P., Haziza, D., Massa, F., El-Nouby, A., Assran, M., Ballas, N., Galuba, W., Howes, R., Huang, P.Y., Li, S.W., Misra, I., Rabbat, M., Sharma, V., Synnaeve, G., Xu, H., Jegou, H., Mairal, J., Labatut, P., Joulin, A., Bojanowski, P.: Dinov2: Learning robust visual features without supervision (2023)

\bibitem{peng2023rwkv}
Peng, B., Alcaide, E., Anthony, Q., Albalak, A., Arcadinho, S., Cao, H., Cheng, X., Chung, M., Grella, M., GV, K.K., et~al.: Rwkv: Reinventing rnns for the transformer era. arXiv preprint arXiv:2305.13048  (2023)

\bibitem{beit2}
Peng, Z., Dong, L., Bao, H., Ye, Q., Wei, F.: Beit v2: Masked image modeling with vector-quantized visual tokenizers (2022)

\bibitem{radford2021learning}
Radford, A., Kim, J.W., Hallacy, C., Ramesh, A., Goh, G., Agarwal, S., Sastry, G., Askell, A., Mishkin, P., Clark, J., Krueger, G., Sutskever, I.: Learning transferable visual models from natural language supervision (2021)

\bibitem{whisper}
Radford, A., Kim, J.W., Xu, T., Brockman, G., McLeavey, C., Sutskever, I.: Robust speech recognition via large-scale weak supervision. In: International Conference on Machine Learning. pp. 28492--28518. PMLR (2023)

\bibitem{rando2022exploringadversarialattacksdefenses}
Rando, J., Naimi, N., Baumann, T., Mathys, M.: Exploring adversarial attacks and defenses in vision transformers trained with dino (2022), \url{https://arxiv.org/abs/2206.06761}

\bibitem{Schlarmann_2023_ICCV}
Schlarmann, C., Hein, M.: On the adversarial robustness of multi-modal foundation models. In: Proceedings of the IEEE/CVF International Conference on Computer Vision (ICCV) Workshops. pp. 3677--3685 (October 2023)

\bibitem{sitawarin2023defending}
Sitawarin, C., Chang, J., Huang, D., Altoyan, W., Wagner, D.: Defending against transfer attacks from public models (2023)

\bibitem{sun2024massive}
Sun, M., Chen, X., Kolter, J.Z., Liu, Z.: Massive activations in large language models (2024)

\bibitem{sun2023retentive}
Sun, Y., Dong, L., Huang, S., Ma, S., Xia, Y., Xue, J., Wang, J., Wei, F.: Retentive network: A successor to transformer for large language models. arXiv preprint arXiv:2307.08621  (2023)

\bibitem{touvron2022deit}
Touvron, H., Cord, M., J{\'e}gou, H.: Deit iii: Revenge of the vit. In: European Conference on Computer Vision. pp. 516--533. Springer (2022)

\bibitem{transformer}
Vaswani, A., Shazeer, N., Parmar, N., Uszkoreit, J., Jones, L., Gomez, A.N., Kaiser, L.u., Polosukhin, I.: Attention is all you need. In: Guyon, I., Luxburg, U.V., Bengio, S., Wallach, H., Fergus, R., Vishwanathan, S., Garnett, R. (eds.) Advances in Neural Information Processing Systems. vol.~30. Curran Associates, Inc. (2017), \url{https://proceedings.neurips.cc/paper_files/paper/2017/file/3f5ee243547dee91fbd053c1c4a845aa-Paper.pdf}

\bibitem{attnsinks}
Xiao, G., Tian, Y., Chen, B., Han, S., Lewis, M.: Efficient streaming language models with attention sinks (2023)

\bibitem{xie2021segformer}
Xie, E., Wang, W., Yu, Z., Anandkumar, A., Alvarez, J.M., Luo, P.: Segformer: Simple and efficient design for semantic segmentation with transformers. Advances in Neural Information Processing Systems  \textbf{34},  12077--12090 (2021)

\bibitem{yuan2022decentralized}
Yuan, B., He, Y., Davis, J., Zhang, T., Dao, T., Chen, B., Liang, P.S., Re, C., Zhang, C.: Decentralized training of foundation models in heterogeneous environments. Advances in Neural Information Processing Systems  \textbf{35},  25464--25477 (2022)

\bibitem{ade20k}
Zhou, B., Zhao, H., Puig, X., Fidler, S., Barriuso, A., Torralba, A.: Scene parsing through ade20k dataset. In: Proceedings of the IEEE Conference on Computer Vision and Pattern Recognition (CVPR) (July 2017)

\bibitem{zhou2024training}
Zhou, J., Chen, Y., Hong, Z., Chen, W., Yu, Y., Zhang, T., Wang, H., Zhang, C., Zheng, Z.: Training and serving system of foundation models: A comprehensive survey. arXiv preprint arXiv:2401.02643  (2024)

\bibitem{ibot}
Zhou, J., Wei, C., Wang, H., Shen, W., Xie, C., Yuille, A., Kong, T.: ibot: Image bert pre-training with online tokenizer (2022)

\bibitem{zhu2024visionmamba}
Zhu, L., Liao, B., Zhang, Q., Wang, X., Liu, W., Wang, X.: Vision mamba: Efficient visual representation learning with bidirectional state space model. arXiv preprint arXiv:2401.09417  (2024)

\end{thebibliography}
\end{document}